# BDSLI: A hybrid CNN-Transformer model for Bengali Sign Language interpretation

Abir Bin Yousuf
Muhammad Iqbal Hossain

**Abstract**

This study introduces a novel hybrid CNN-Transformer architecture to address the limited progress in Bengali SLR, focusing on isolated sign word recognition and sentence generation. This specific model combination is new to Bengali SLR tasks. A custom video dataset was developed, featuring 62 distinct Bengali sign words (250 samples/class), along with a separate test dataset. The CNN-Transformer model demonstrated superior performance against all comparative and baseline models (e.g., CNN-LSTM, standalone TCN), achieving a 99.58% training accuracy (99.48% validation) and a 98.65% test accuracy. The trained model was subsequently deployed in a web application for real-world validation.

# 1 Introduction

According to the World Federation of the Deaf, sign language is the main means of communication for more than 70 million people around the world [1]. However, most people do not have the time or resources to learn a new language from scratch. To solve this problem, researchers have been developing sign language recognition (SLR) systems to facilitate communication with the deaf and mute community. Although languages like ASL and BSL have seen significant progress in SLR, Bengali Sign Language (BDSL) remains underexplored.

After observing that, we decided to focus on building a real-time system to recognize Bengali sign words and generate meaningful sentences. The ultimate goal of this study is to enable natural face-to-face conversations between hearing individuals and members of the Bengali deaf community with the help of a smartphone or Web-based application.

In reviewing previous SLR and sign language translation (SLT) research, we found that models such as HMMs, LSTM, GRU, and CNN+LSTM have been widely used, often trained on image datasets. In contrast, we propose a video-based hybrid CNN–Transformer model, leveraging CNN's ability to capture local patterns and Transformers' strength in modeling global dependencies. Due to the lack of publicly available BDSL video datasets, we created our own dataset with 62 different Bengali words and trained the hybrid model and seven other popular models on it. Our model achieved the highest accuracy on both training and unseen test datasets. Finally, we deployed the model in a real-time web application. This app can interpret Bengali sign words through a webcam and generate complete sentences with the interpreted words.

This research paper is organized as follows: Section 2 reviews related work on sign language recognition. Section 3 provides a detailed description of the datasets and the feature extraction technique. Section 4 presents the architecture of the proposed hybrid model along with the training procedure. In Section 5, the experimental results are analyzed and compared with several standalone and hybrid benchmark models. Finally, Section 6 concludes the paper and highlights possible directions for future work.

# 2 Related Work

When we were reviewing related research work, we noticed that BDSL is a sign language that is comparatively less explored. There are some studies on BDSL, but they do not work with sentence generation. In the research paper [2], the authors managed to detect isolated sign words from input videos. For this study, they used YOLOV5 and PyTorch to build their model and trained the model on a dataset with 34 different words. The trained model managed to achieve a training accuracy of 76. 29%. However, this model is not suitable for dynamic gestures. Therefore, we decided to use CNN after reviewing the research papers by Zhang et al. [3], Ramar et al. [4], and Halvardsson et al. [5]. These studies showed that CNN can perform extremely well in this field of research. Unfortunately, all of these papers worked with alphabet recognition and the models trained for these papers were unable to interpret sign words.

A notable research paper by Kumari et al. [6] reports using a hybrid CNN+LSTM model combined with MobileNetV2. The authors trained their model on the WLASL benchmark dataset. This suggested architecture achieved an accuracy of 84.65%, which is a notable achievement. However, the authors worked with word-level recognition and their system was not programmed to construct sentences with the predicted words. In another paper, Chaikaew et al. [7] claimed that the model they proposed based on Recurrent Neural Networks (RNN) can accurately recognize sign words just like other popular models such as LSTM, BiLSTM and GRU. In their paper, they worked with Thai Sign Language (TSL) and used mediapipe to extract hand landmarks from the dataset. We took inspiration from this research paper to use mediapipe, while trying to considerably increase the accuracy.

The research paper by Paul et al. [8] discusses several approaches for sign language recognition including a ResNet-based CNN model, LSTM and GRU. However, these models were only trained on 3 words and 26 alphabets. Hence, generating full-length meaningful sentences was not possible for them. A different approach was proposed in the research work of Ma et al. [9], where the authors introduced four different two-stream mixed CNN models and demonstrated that TSM-ResNet50 achieved the best performance. The researchers achieved 97.57% accuracy by training the TSM-ResNet50 model separately on the MNIST and ASL datasets.

In the research paper by Sreemathy et al. [9], two different models were used for word-level SLR. In their paper, they trained SVM and YOLOV4 on a dataset with 80 different word classes. Despite having a large number of classes, the models would not support dynamic gestures as they were trained on static images of words. Lastly, Rahman et al. [10] proposed Generative Adversarial Networks (GANs) for word-level SLR. In their paper, they have shown that their model can achieve 91.3% accuracy after being trained on a dataset with 20 ASL words.

In this study, we are trying to surpass the accuracy achieved in all of these previous papers and overcome their limitations with our hybrid CNN + Transformers model.

# 3 Data Preparation and Feature Extraction

## 3.1 Dataset Description

At first, we needed to determine the specific words we wanted to use for our dataset. From hundreds of words, it was narrowed down to 62 words that are often used in our daily conversations. Figure 1 shows the selected Bengali sign words that were used to create the datasets for this study. As mentioned earlier, two distinct datasets were constructed for this study: one to train the models and the other to evaluate their performance on unseen data.

আজ, আবার, আমাদের, আমার, আমি, আসা, কবে, কয়টা বাজে,
কাজ, কিভাবে, কী, কে, কেন, কোথায়, খাওয়া, ছেলে, ডাকা, তাদের,
তুমি, তোমাদের, দয়া করে, দুঃখিত, দুপুর, দেখতে, দেখা, দেখা হয়ে,
দেশ, ধন্যবাদ, না, নাম, পড়া, বন্ধু, বলা, বাংলা, বাড়ী, বাবা, বুধবার,
বৃহস্পতিবার, বোন, ভাই, ভালো, ভাষা, মঙ্গলবার, মা, মেয়ে, যাওয়া,
রবিবার, রাত, লেখা, শনিবার, শুক্রবার, শোয়া, সকাল, সাহায্য, সিনেমা,
সুন্দর, সে, সোমবার, স্ত্রী, স্বামী, হবে, হ্যা

*Figure 1: Selected Bengali sign words for dataset construction*

**Training Dataset:** For the training dataset, we collected 250 videos for each word and 15500 videos in total. The sequence length of each video was 30, indicating that a video had 30 frames in it. We tried to increase the number of frames to 60; however, it would have reduced the accuracy since we were working with real-time prediction. Hence, it was agreed to keep the sequence length at 30. Figure 2 shows the screenshots taken from the sample videos of the dataset. The top left image in Figure 2 illustrates the sign for the word ’Ami’, while the top right image shows the sign for ‘Dekha’. The remaining pictures represent the Bengali words ’Baari’, ’Ha’, ’Shokal’ and ’Shahajjo’, respectively.

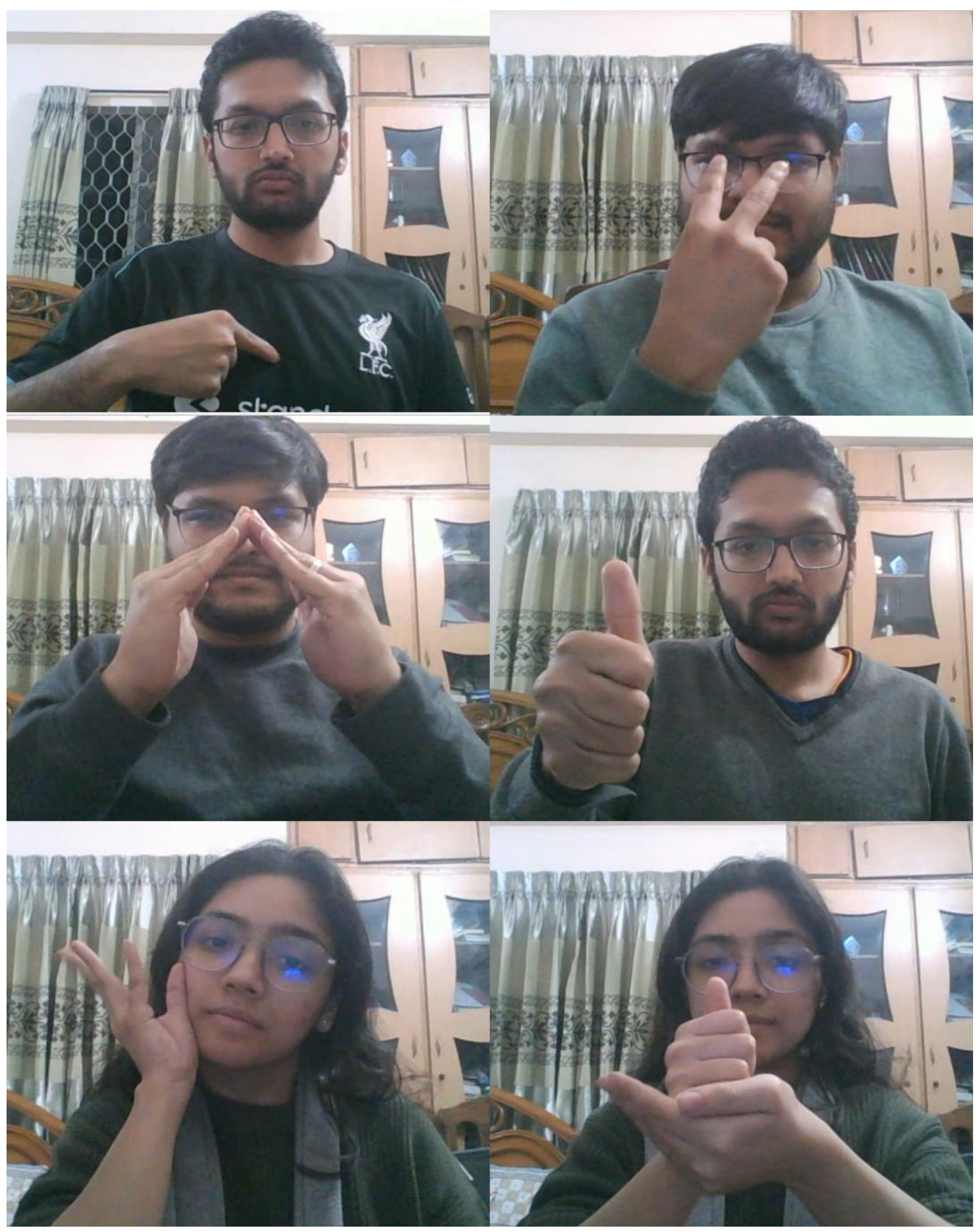

*Figure 2: Sample video frames of Bengali sign words*

**Test Dataset:** For constructing the test dataset, the videos were captured again so that we could obtain a dataset that is different from the training dataset. This is to ensure that there is some unseen data to find out whether the model can predict Bengali sign words from these unseen videos. Here, a total of 3100 videos were captured, with 50 videos per word. The sequence length was kept at 30, just like the training dataset.

## 3.2 Feature Extraction Technique

**Hand Landmarks with Relative Position:** As we were working with sign language, where hand gestures are the most important things, we decided to extract hand landmarks from our dataset. A Python library called mediapipe was used to extract hand landmarks from each video. The hand landmarks show us the posture of the hands, and the landmarks are drawn differently for non-identical gestures. Figure 3 illustrates examples of hand landmarks extracted for selected Bengali sign words.

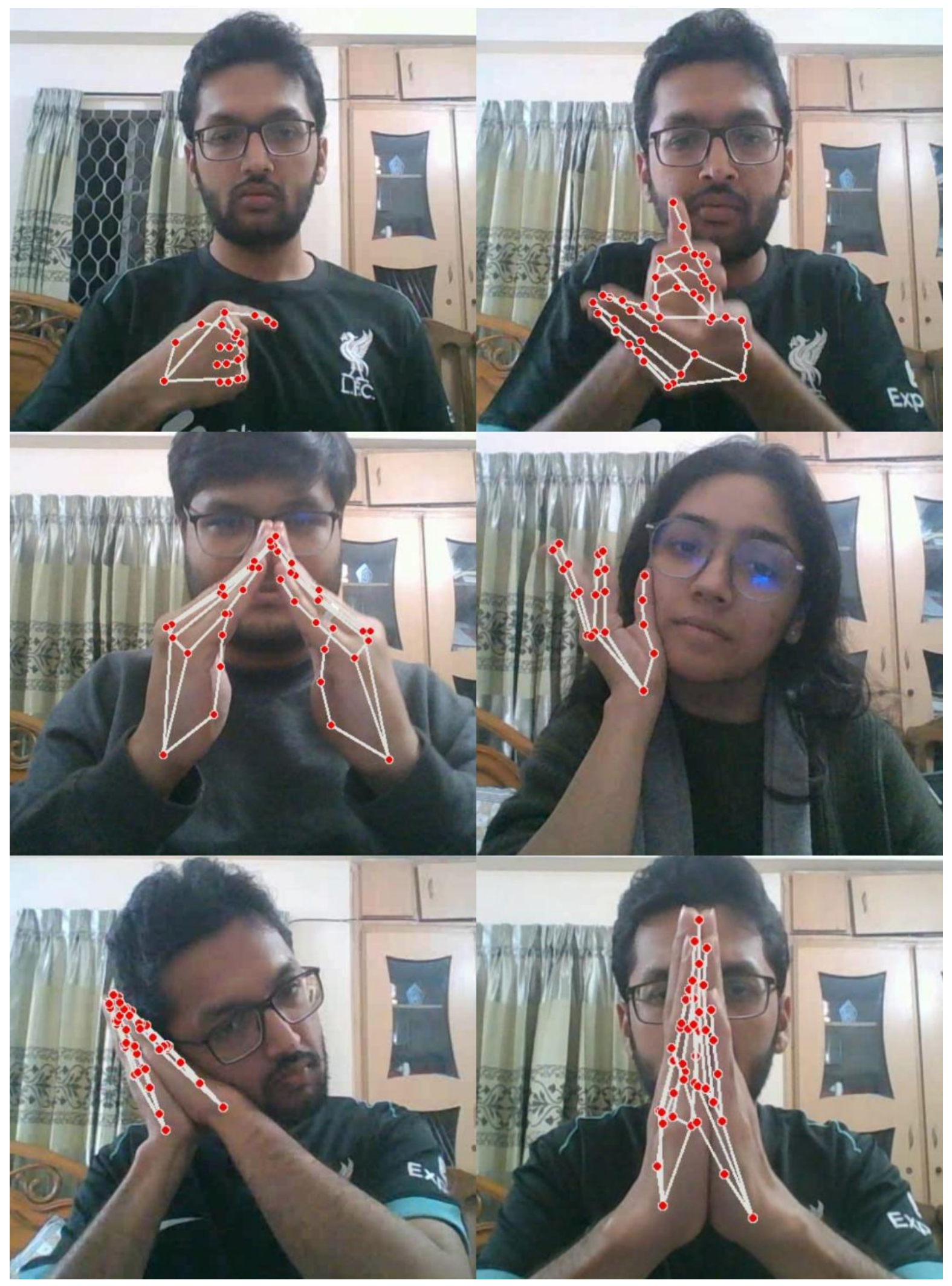

*Figure 3: Extracted hand landmarks for the words 'Ami', 'Shahajjo', 'Baari', 'Shokal', 'Showa' and 'Doya Kore'*

Here, each hand consists of 21 landmarks and each landmark provides three spatial coordinates: $(x, y, z)$. The extracted features consist of:

1. **Extracted Landmark Coordinates:** Each detected hand provides 21 landmarks, and each landmark $i$ has three coordinates $(x_i, y_i, z_i)$. The feature vector of landmarks is represented as:

$$F_{\text{landmarks}} = \{(x_1, y_1, z_1), (x_2, y_2, z_2), \ldots, (x_{21}, y_{21}, z_{21})\}$$

   This results in a feature vector of size $21 \times 3 = 63$.

2. **Hand Position Calculation:** To capture the overall position of the hand, we compute the centroid of the 21 landmarks using the mean of their coordinates:

$$x_{\text{avg}} = \frac{1}{N}\sum_{i=1}^{N} x_i, \quad y_{\text{avg}} = \frac{1}{N}\sum_{i=1}^{N} y_i, \quad z_{\text{avg}} = \frac{1}{N}\sum_{i=1}^{N} z_i$$

In the provided equation, $N$ represents the total number of hand landmarks and $N = 21$, as stated in the book by Bhatti et al. (**bhatti2024deep?**). These three values $(x_{\text{avg}}, y_{\text{avg}}, z_{\text{avg}})$ are added to the feature vector.

3. **Final Feature Vector:** The final feature vector for a single frame consists of the landmark coordinates along with the average hand position:

$$F = [x_1, y_1, z_1, x_2, y_2, z_2, \dots, x_{\text{avg}}, y_{\text{avg}}, z_{\text{avg}}]$$

This results in a total of $63 + 3 = 66$ features per frame.

In the final feature vector, 3 is added to the total number of features. This represents the relative position of the hands. A closer examination of the lower section of Figure 3 reveals that the signs for 'Showa' and 'Doya kore' exhibit identical hand gestures. So, when the hand landmarks are extracted, they also look almost the same. These identical words resulted in the model making wrong predictions from time to time. The additional feature (relative position) was introduced to solve this issue. As shown in Figure 3, the position of the hands for 'Showa' is slightly bent to the left, while the hands are exactly in the middle of the frame for 'Doya kore'. This positional difference was captured as a new feature to help the model distinguish between otherwise similar gestures.
**Feature Extraction Time:** The extraction of hand landmarks along with their relative positions took approximately 4.25 hours for the training dataset and approximately 1.25 hours for the test dataset. The extracted features from these datasets were saved in two separate pickle files.

# 4 Model Description and Experimental Results

## 4.1 Model Architecture

The proposed hybrid CNN+Transformers model is designed to effectively recognize Bengali Sign Language gestures by combining convolutional feature extraction (CNN) and self-attention mechanisms (Transformers). Figure 4 illustrates the entire workflow of the hybrid model using a flow chart and marks the 4 key components using different colors.

### 4.1.1 Input Layer

The model takes a sequence of preprocessed feature vectors, which were extracted from Bengali Sign Language videos, as input. Each input sequence is represented as:

$$X \in \mathbb{R}^{T \times F}$$

where $T$ represents the length of the sequence (number of time steps) and $F$ is the number of features extracted per time step. The input shape for the model is:

(sequence_length,num_features)

This input shape ensures that the model can process sequential data effectively.

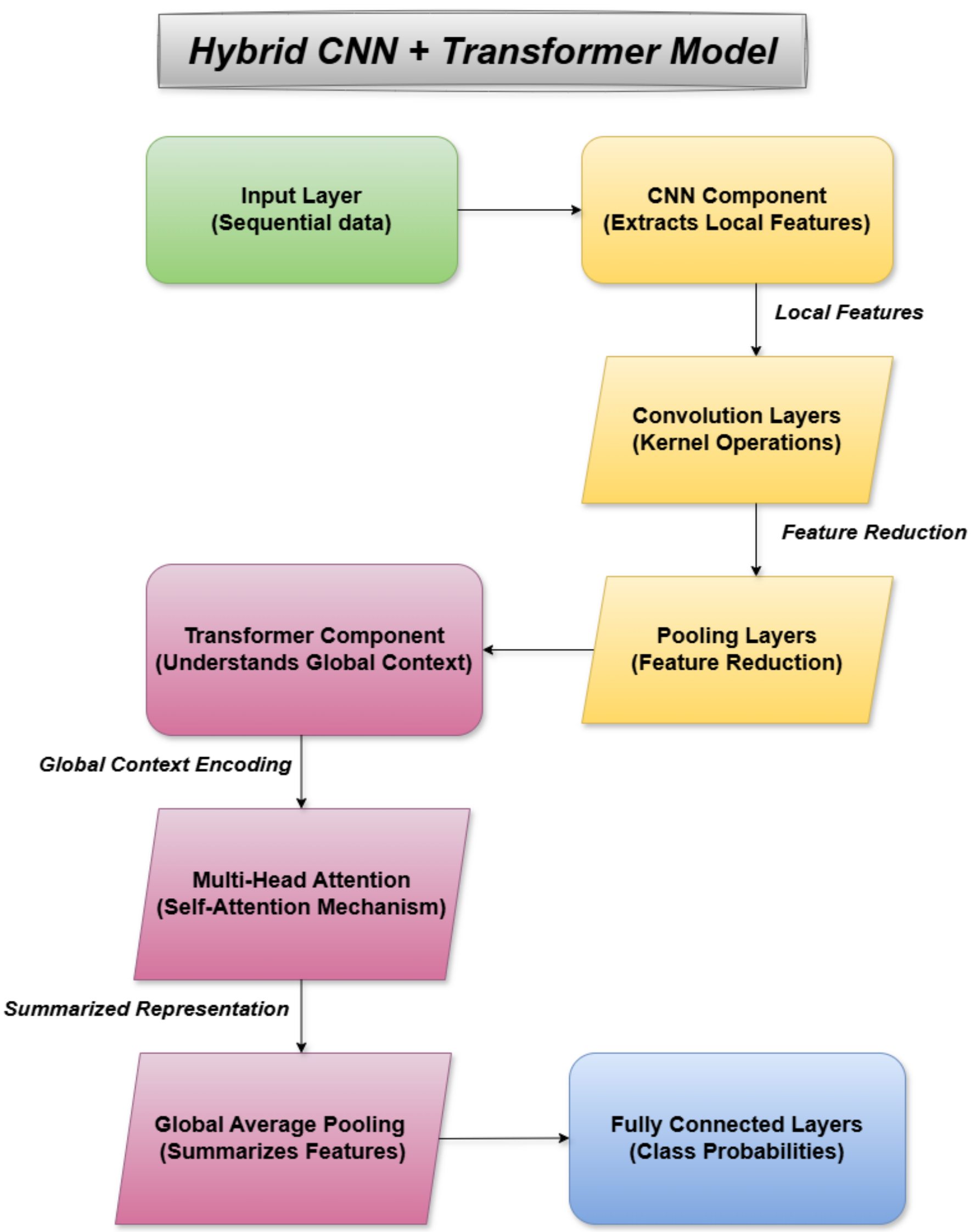

*Figure 4: The complete workflow of the proposed hybrid CNN+Transformer model*

### 4.1.2 CNN-Based Feature Extraction

To extract spatial and local temporal features from the input sequence, the model uses Conv1D (1D Convolutional Neural Networks) as described in the book by Aggarwal et al. (**aggarwal2025integrated?**). CNN is well suited for capturing patterns within local regions of the sequence, such as hand movements or shape variations over short periods. The feature extraction process involves the following layers:

- **First Convolutional Layer (Conv1D):** This layer uses 64 filters with a kernel size of 3. It applies the ReLU activation function and uses ’same’ padding, which ensures that the output size remains unchanged. The layer is designed to capture low-level spatial and temporal patterns from the input sequence.

- **Max-Pooling Layer (MaxPooling1D):** This layer uses a pool size of 2 to reduce the sequence length by half. In this way, the most important features are effectively retained while decreasing computational complexity.

- **Second Convolutional Layer (Conv1D):** This layer consists of 128 filters with a kernel size of 3, using the ReLU activation function and 'same' padding. This layer captures deeper hierarchical features and is capable of detecting more complex movement patterns in the input data.

- **Max-Pooling Layer (MaxPooling1D):** This layer uses a pool size of 2 to further reduce dimensionality. This, in turn, enables the model to focus on the most salient features.

- **Dropout Layer:** This layer applies a dropout rate of 0.3 to avoid overfitting by randomly dropping some neurons during training. It helps to make the model more robust.

At this stage, the input sequence has been transformed into a set of high-level feature representations that serve as the input for the Transformer-based attention mechanism.

### 4.1.3 Transformer-Based Attention Mechanism

Although CNN is effective in feature extraction, it struggles in capturing long-range dependencies in sequential data. To address this limitation, the model integrates a self-attention mechanism inspired by Transformers. The attention mechanism consists of the following layers:

- **Layer Normalization:** This layer helps stabilize activations and gradients by normalizing feature distributions. As a result, faster and more stable training is ensured.

- **Multi-Head Self-Attention (MHSA):** This layer uses 4 attention heads with a key dimension of 128. Each attention head learns a unique representation of temporal relationships within the feature sequence by computing scaled dot-product attention to focus on the most relevant time steps. According to the research paper by Vaswani et al. (**vaswani2023attentionneed?**), the self-attention mechanism is formally defined as:

  $$\text{Attention}(Q, K, V) = \text{softmax}\left(\frac{QK^T}{\sqrt{d_k}}\right)V$$

  where:

  - $Q$ (Query), $K$ (Key), and $V$ (Value) are projections of the input sequence.
  - $d_k$ is the dimension of the key vectors used for scaling.

- **Global Average Pooling (GAP):** This layer computes the average of the attention outputs across time steps, reducing the dimensionality of the sequence while retaining crucial information. By replacing fully connected layers with a parameter-free operation, it also helps prevent overfitting.

By leveraging Multi-Head Self-Attention, the model learns to highlight the most significant temporal features of a given sign. In this way, classification performance is considerably improved.

### 4.1.4 Fully Connected Layers (Feedforward Network)

After extracting both local (CNN) and long-range (Transformers) features, the processed feature representations are passed to fully connected layers so that these layers can classify different class labels. The final layers contain the following components:

- **Dense Layer (Fully Connected Layer):** This layer contains 128 neurons with a ReLU activation function. This dense layer also performs high-level feature abstraction by mapping the extracted features to meaningful representations.

- **Dropout Layer:** This layer applies a dropout rate of 0.3 to avoid overfitting by randomly deactivating neurons during training, which enhances the generalization ability of the model.

- **Output Layer:** This layer uses a softmax activation function and contains several neurons equal to the number of classes (62) in the dataset. This layer also converts the final feature representations into class probabilities.

## 4.2 Model Training and Evaluation

As mentioned above, the hand landmarks along with their relative positions were saved in a pickle file, which was subsequently used to train our model. To begin, essential Python libraries such as numpy, pickle, tensorflow, matplotlib.pyplot and sklearn were imported. The remaining workflow can be divided into the following parts:

1. **Data Preprocessing:** In this step, data are loaded from a pickle file which contains features (X), labels (y) and class names (class names). As mentioned in the book by Bhatti et al. [12], the features (X) are usually normalized by dividing them by the maximum value. This ensures that the input data are appropriately scaled for the neural network. The training dataset was split into training and validation sets using an 80-20 ratio with the help of the Python library scikit-learn.

2. **Compilation and Training:** As there are multiple word classes, we decided to train the hybrid model with Adam optimizer and chose sparse categorical cross-entropy as the loss function. The model was trained for 100 epochs, and the training process included monitoring the validation accuracy. The model was trained with a batch size of 32 and a learning rate of 0.0001, which produced the best results during experimentation. Among the different epoch settings tested, 100 epochs achieved the highest accuracy.

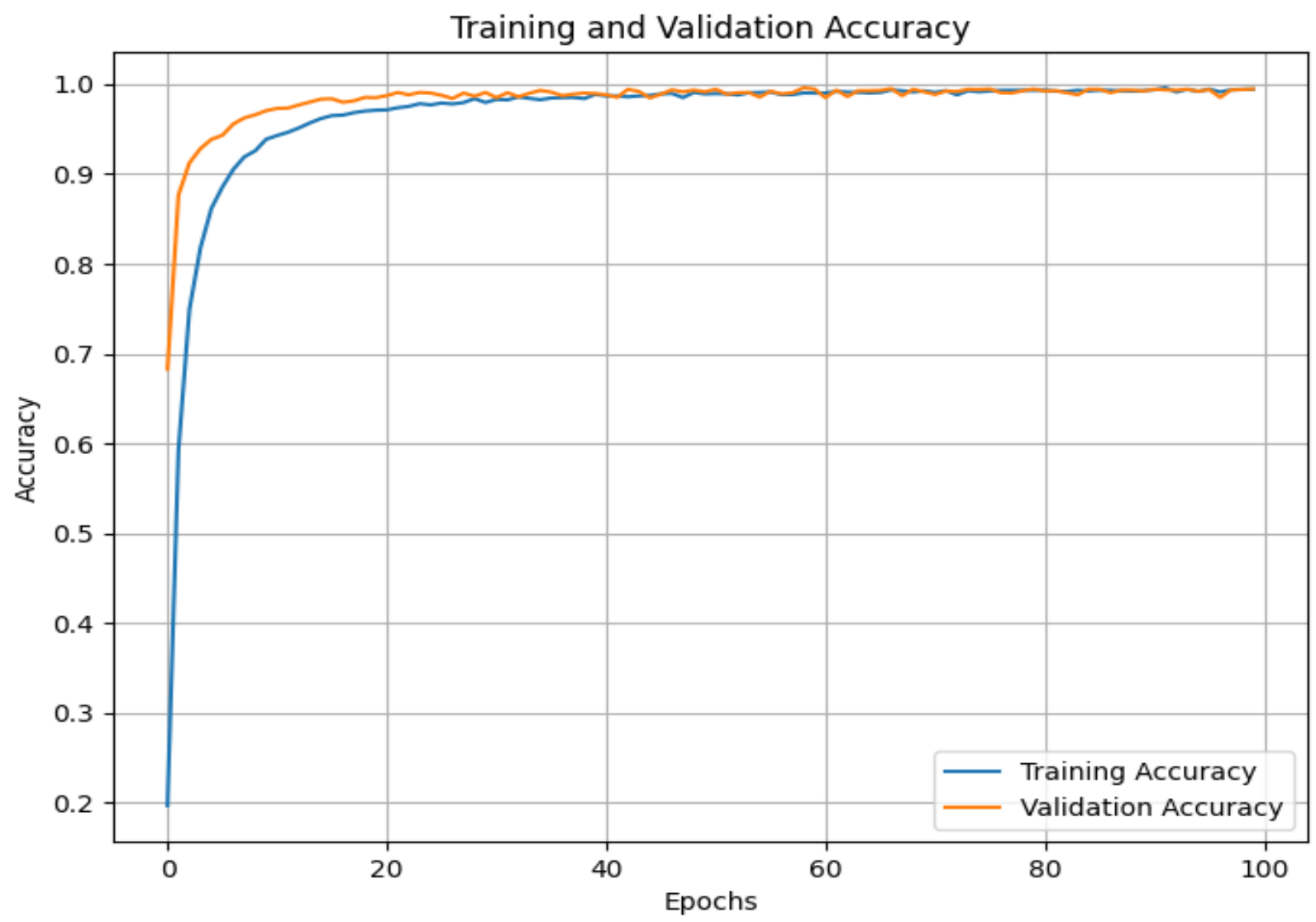

*Figure 5: Training and Validation accuracy of the Hybrid Model*

3. **Model Evaluation:** Figure 5 shows the training and validation accuracy achieved by the model throughout 100 epochs. The highest training accuracy (99.58%) was recorded for the final epoch with only 1.27% training loss. In the final epoch, the model also achieved a validation accuracy of 99.48% with a validation loss of 2.07%. The highest values of training accuracy and validation accuracy were almost the same, indicating that the model had converged perfectly during training.

   Nevertheless, further evidence was still needed to prove that the model was indeed performing well. So, the precision, recall, and F1 scores were printed after each epoch. Figure 6 illustrates the gradual improvement of these values over the course of 100 epochs. The highest scores were obtained on the final epoch where precision was 99.50, recall was 99.48 and F1 score was 99.48. The scores suggest that the model performs excellently and consistently, with very few misclassifications.

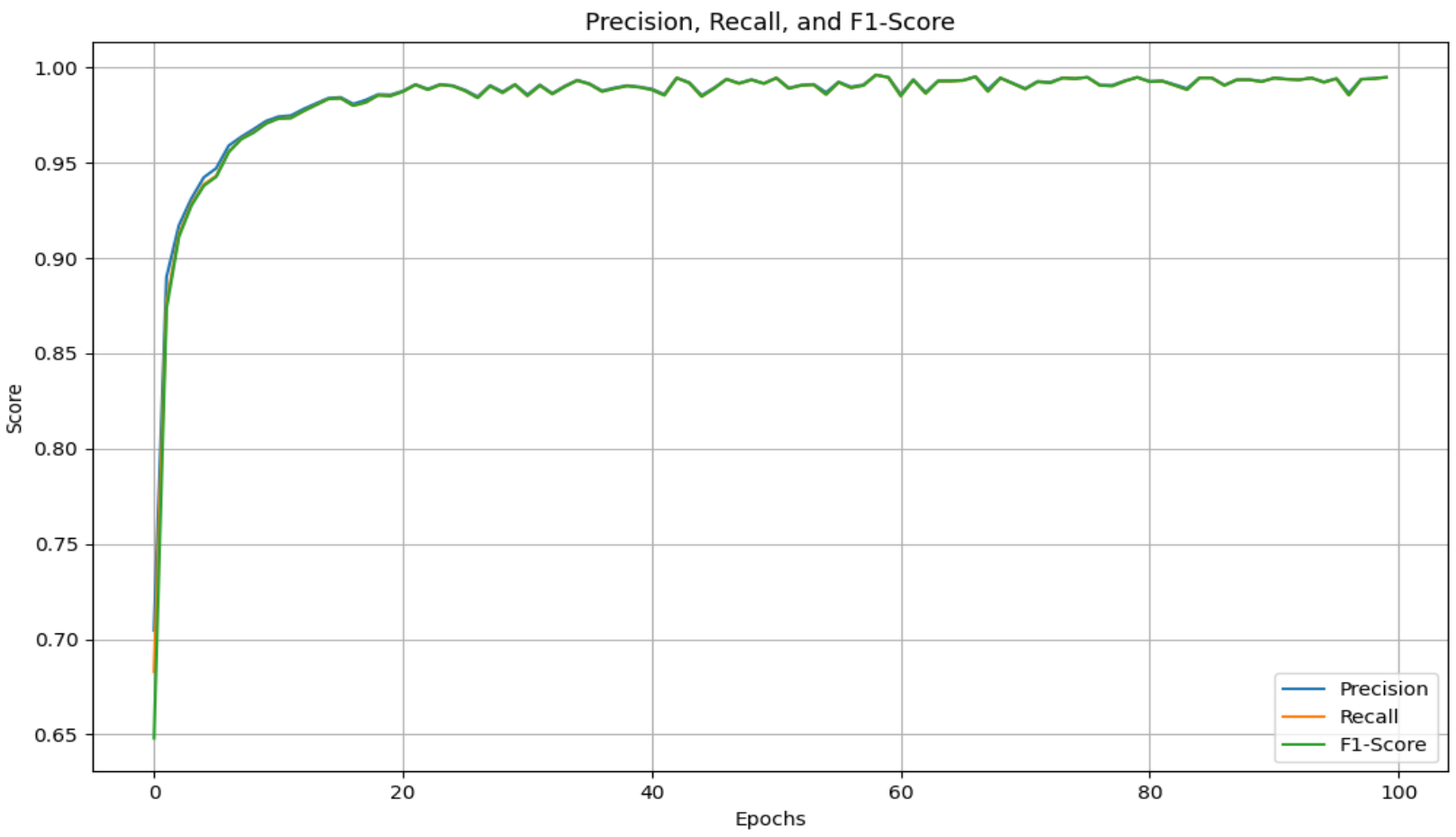

*Figure 6:Precision, Recall and F1 score of the Proposed Hybrid Model*

**Training Time:** The model completed the training process in a short amount of time. It only took the model around 19 minutes and 57 seconds to complete 100 epochs. This training time was notably efficient compared to the other models trained in this study.

# 5 Comparative Analysis of Model Performance

In this study, seven additional models were trained to demonstrate the effectiveness of the proposed approach. All models were trained on the same dataset under identical experimental settings to ensure a fair comparison.

## 5.1 Comparison of Training Performance

After training the models, they were compared using a combined table, where the first evaluation on the table presents the training and validation accuracy along with the total training time. The second evaluation on the table will show and compare the precision, recall, and F1 scores that were recorded for the trained models.

From Table 1, it is evident that our model achieved the highest training and validation accuracy, which are 99.58% and 99.48%. CNN+LSTM obtained the second highest training accuracy (99.37%), and the second highest validation accuracy (99.35%) was achieved by Transformers. GRU model had the worst performance on this list with a training accuracy of 97.88% and a validation accuracy of only 95.94%. This model took almost 48 minutes to train, which is second to the longest time (53m 36s) taken by Transformers.

*Table 1: Training accuracy, Validation accuracy and Training Time with their respective models*

| Model | Train acc | Val acc | Training Time |
|---|---|---|---|
| *CNN+Transformers* | **99.58%** | **99.48%** | 19m 56s |
| *CNN+LSTM* | 99.37% | 98.65% | 17m 46s |
| *GRU* | 97.88% | 95.94% | 47m 39s |
| *LSTM* | 99.36% | 97.61% | 37m 4s |
| *TCN* | 99.29% | 98.94% | 28m 15s |
| *Transformers* | 99.34% | 99.35% | 53m 36s |
| *CNN+BiLSTM* | 99.32% | 99.00% | 18m 57s |
| *CNN+GRU* | 98.39% | 97.97% | **12m 29s** |

However, when we combined GRU with CNN, the performance improved significantly and this combination took the lowest amount of time (12m 29s) to complete training. Although CNN + GRU had the lowest training time, the training and validation accuracy were still lower than most models. Our model (CNN + Transformers) took approximately 20 minutes to train, which is impressive compared to other models.

*Table 2: Additional Performance Metrics to Compare the models*

| Model | Precision | Recall | F1 Score |
|---|---|---|---|
| *CNN+Transformers* | **99.50** | **99.48** | **99.48** |
| *CNN+LSTM* | 98.67 | 98.65 | 98.64 |
| *GRU* | 96.13 | 95.94 | 95.92 |
| *LSTM* | 97.74 | 97.61 | 97.61 |
| *TCN* | 99.03 | 98.94 | 98.95 |
| *Transformers* | 99.41 | 99.35 | 99.36 |
| *CNN+BiLSTM* | 99.04 | 99.00 | 99.00 |
| *CNN+GRU* | 98.06 | 97.97 | 97.96 |

In Table 2, the precision, recall, and F1 scores are presented and the highest values are marked using bold font. It is apparent that CNN+Transformer has the highest values, which are 99.50, 99.48 and 99.48. Transformers takes the second place with the values 99.41, 99.35 and 99.36. As anticipated, the GRU model has the lowest scores, while the LSTM model performs slightly better. Another important thing to note here is that both TCN and CNN+BiLSTM have better values than CNN+LSTM even though CNN+LSTM had higher training accuracy.

## 5.2 Performance Comparison on Test Dataset

As we have mentioned earlier, a separate test dataset was prepared to evaluate the trained models on unseen videos. Figure 7 demonstrates the performance of each model on the test dataset.

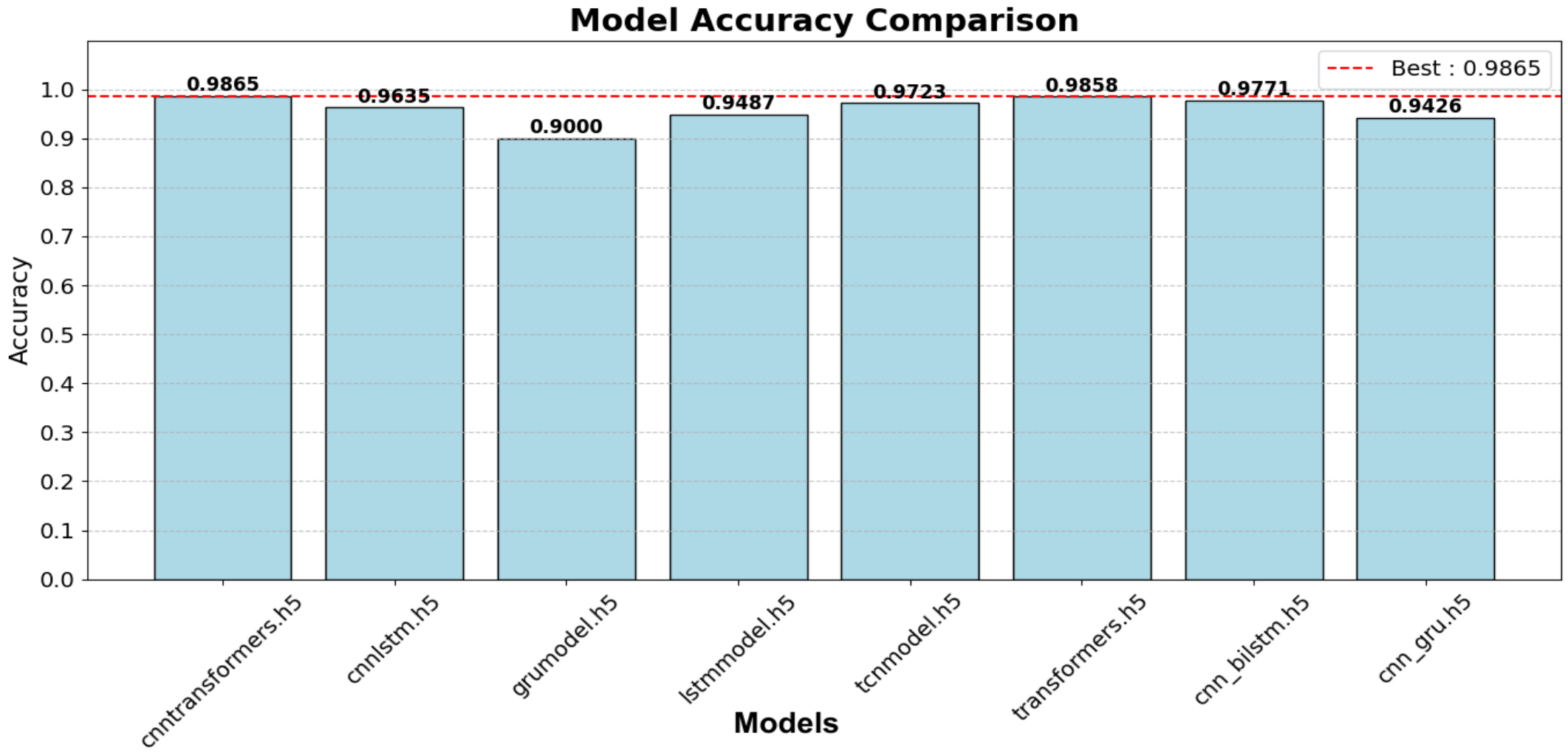

*Figure 7: Performance displayed by the models on Test Dataset*

In Figure 7, the highest accuracy is 98.65%, which was achieved by the proposed hybrid model. The second and third highest accuracies (98.58% and 97.71%) were obtained by Transformers and CNN+BiLSTM respectively. In contrast, the GRU model has the lowest test accuracy of 90.00%. The hybrid CNN+GRU model has the second lowest accuracy (94.26%) for the test data. These percentages indicate that the proposed hybrid model outperforms other models, with Transformers being the only model that can closely match its performance.

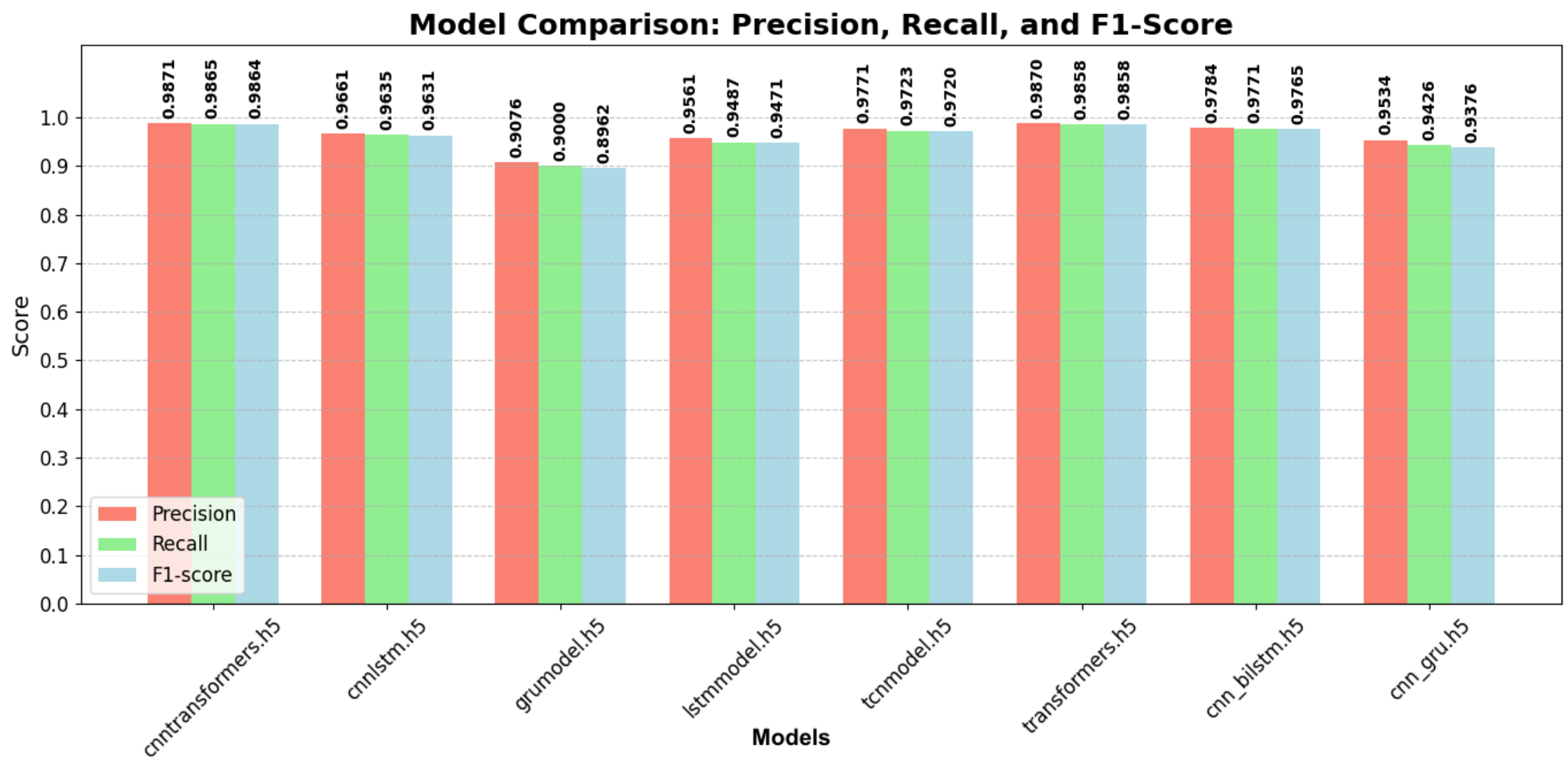

*Figure 8: 3 Different Performance Metrics to Evaluate the Models on Test data*

For further assessment, the test dataset was used to analyze the precision value, recall value and F1 score for each model. Figure 8 presents a comparative bar graph summarizing these metrics for all models. The highest values were once again recorded for the proposed hybrid model, which are 98.71, 98.65 and 98.64, respectively. Transformers takes the second place with the values 98.70, 98.58 and 98.58, whereas GRU model has the lowest values (90.76, 90.00, 89.62) among the models.

## 5.3 Comparison with Prior Research

In this section, we present a comparative analysis between our model and those proposed by other researchers, as cited in Section 2. Table 3 compares the models with multiple evaluation metrics to assess their effectiveness in the SLR task.

*Table 3: Comparison between the hybrid model and the models proposed in earlier research papers*

| **Paper** | **Model** | **Classes** | **Word** | **D-G** | **Acc.** |
|---|---|---|---|---|---|
| *Tazalli (**tazalli2022computer?**)* | YOLOV5 | 34 | | | 76.29% |
| *Zhang (**Zhang2024?**)* | DPCNN | 24 | | | 99.52% |
| *Gayathri (**D._Ramar_2024?**)* | Mobile-Net V2 | 26 | | | 85.45% |
| *Halvardsson (**halvardsson2020?**)* | InceptionV3 | 26 | | | 85% |
| *Kumari (**kumari2024isolated?**)* | CNN+LSTM | 100 | | | 84.65% |
| *Paul (**paul2024adam?**)* | CNN+LSTM | 29 | | | 94.3% |
| *Ma (**ma2022two?**)* | TSM-ResNet50 | 29 | | | 97.57% |
| *R.Sreemathy (**sreemathy2023continuous?**)* | YOLOV4 | 80 | | | 98.8% |
| *Rahman (**rahman2021word?**)* | GANs | 20 | | | 91.3% |
| ***Proposed Model*** | **CNN+T** | 62 | | | **99.58%** |

In Table 3, the column labeled 'Word' indicates whether a model supports word-level sign language recognition. The abbreviation D-G stands for Dynamic Gestures, indicating whether a model can recognize dynamic gestures. Most of the papers listed here do not address full-length sentence generation, whereas our study specifically covers this aspect. Table 3 also shows that the proposed model achieves the highest accuracy and effectively recognizes sign words that involve dynamic gestures.

Here, YOLO was used as the primary model in the research papers by Tazalli et al. [2] and Sreemathy et al. [10]. It is known that YOLO is trained on image datasets and as a result, this model does not support dynamic hand movements. In contrast, the proposed approach supports dynamic gestures as it was trained on a video dataset. The chosen hybrid model also has better accuracy than YOLO.

In their research experiments, Paul et al. [8] and Ma et al. [9] used datasets with 29 different classes. However, they only had 3 words in their datasets and the remaining classes were English alphabets. On the other hand, we did not include alphabets in our dataset, and it has 62 distinct Bengali sign words. Lastly, the research paper by Kumari et al. [6] mentions working with more word classes (100), where the dataset also included words with dynamic gestures. We have trained the same model proposed in their study with our BDSL dataset and found that our CNN+Transformers can outperform their model for both training and test datasets.

# 6 Conclusion and Future Work

This research aimed to develop a robust hybrid model for Bengali Sign Language Recognition (SLR) and integrate the model into an Android application. Such an app can make communication with the deaf community a lot easier. A CNN-Transformer hybrid model was trained on a custom dataset of 62 Bengali words and evaluated against several existing models.

Due to the time and resource demands of extensive experimentation, the model was integrated into a lightweight web-based application rather than a full-fledged Android app. In addition, the lack of natural language processing (NLP) expertise limited the scope of deeper language integration. Furthermore, the model occasionally misclassifies sign words that are visually identical. All of these constraints underscore the need for further technical development and collaboration.

Our future studies will focus on building a bidirectional system that is capable of both recognizing and generating sign language. We will try to develop an Android app where input sentences are translated into coherent animated sign sequences. We believe that this complex vision can come true by working in collaboration with an NLP expert and an Android app developer.